\documentclass{article}
\usepackage{spconf,amsmath,graphicx}
\usepackage{multirow}
\usepackage{color,soul} 

\title{A Reference-less Quality Metric for Automatic Speech Recognition via Contrastive-Learning of a Multi-Language Model with Self-Supervision}
\name{Kamer Ali Yuksel,
      Thiago Ferreira, 
      Ahmet Gunduz, 
      Mohamed Al-Badrashiny, 
      Golara Javadi}
\address{aiXplain Inc., Los Gatos, CA, USA}

\begin{document}
%
\maketitle
\begin{abstract}
The common standard for quality evaluation of automatic speech recognition (ASR) systems is reference-based metrics such as the Word Error Rate (WER), computed using manual ground-truth transcriptions that are time-consuming and expensive to obtain. This work proposes a multi-language referenceless quality metric, which allows comparing the performance of different ASR models on a speech dataset without ground truth transcriptions. To estimate the quality of ASR hypotheses, a pre-trained language model (LM) is fine-tuned with contrastive learning in a self-supervised learning manner. In experiments conducted on several unseen test datasets consisting of outputs from top commercial ASR engines in various languages, the proposed referenceless metric obtains a much higher correlation with WER scores and their ranks than the perplexity metric from the state-of-art multi-lingual LM in all experiments, and also reduces WER by more than $7\%$ when used for ensembling hypotheses. The fine-tuned model and experiments are made available for the reproducibility: https://github.com/aixplain/NoRefER
\end{abstract}

\section{Introduction}
\label{sec:intro}
Automatic speech recognition (ASR) is a rapidly evolving field that has been actively researched for over six decades. ASR systems have numerous practical applications, including voice assistants, dictation software, and call centers. ASR has become an essential technology for many businesses and individuals, allowing for hands-free interaction and translating spoken language into text. Traditionally, the evaluation of ASR systems has been based on comparing the system's outputs with ground-truth transcripts, also known as references. Reference-based metrics, such as the Word-Error-Rate (WER), are calculated by comparing the outputs of the ASR system with the ground-truth transcripts and determining the number of errors made. The most significant limitation of these metrics is that they require ground-truth transcripts, which may not always be available, and the quality of the reference transcript can affect the accuracy of the evaluation. Instead, referenceless evaluation metrics for ASR might use the audio and output features to estimate the resulting quality. 

\begin{figure}[t]
\centering 
\centerline{\includegraphics[width=\columnwidth]{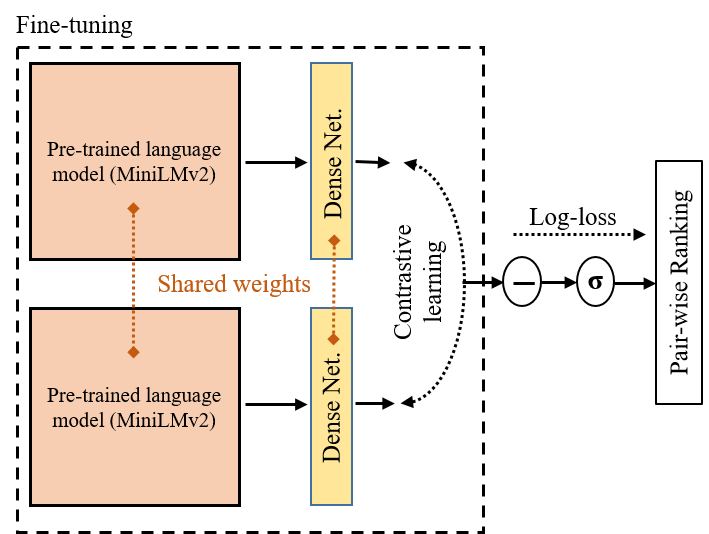}}
\caption{NoRefER fine-tunes a pre-trained language model in a self-supervised learning manner with contrastive learning.}
\label{fig:frame}
\end{figure}

\begin{figure*}[ht]
\centering 
\centerline{\includegraphics[width=\textwidth]{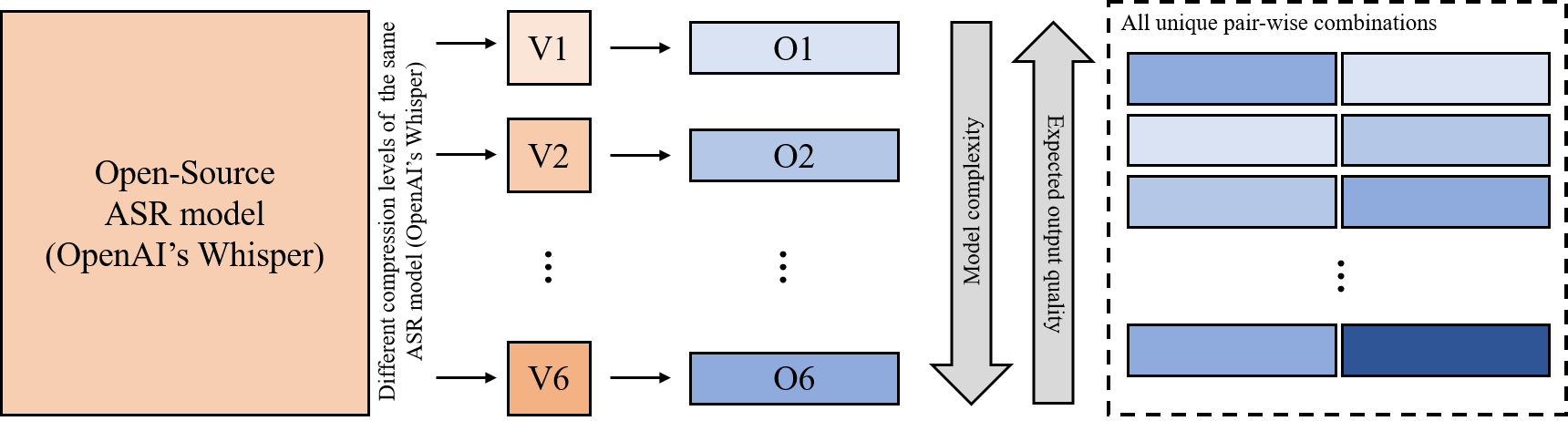}}
\caption{Training and validation dataset generation process. To generate a referenceless set of negative and positive pairs for self-supervised contrastive learning, six compression levels of OpenAI's Whisper~\cite{radford2022robust} is used as ASR models ($V_{1-6}$) to compute different quality outcomes ($O_{1-6}$). All pair-wise combinations of these outputs formed the training and validation dataset.}
\label{fig:dataset}
\end{figure*}

Recently, many efforts have been made to train regression or ordinal classification models for ASR quality estimation based on supervised learning of speech and language features~\cite{del2018speaker,swarup2019improving,jiang2005confidence,kalgaonkar2015estimating}. Fan \textit{et al.} \cite{fan2019neural} proposed using a bidirectional transformer language model conditional on speech features for ASR quality estimation. They designed a neural zero-inflated Beta regression layer, which closely models the empirical distribution of WER, and reported results in WER prediction using the metrics of Pearson correlation and mean absolute error (MAE). Ali and Renals \cite{ali2020word} used a multistream end-to-end architecture with acoustic, lexical, and phonotactic features for estimating WER without having access to the ASR system. In another study, Sheshadri \textit{et al.}~\cite{sheshadri2021bert} proposed a BERT-based architecture with speech features for estimating WER through balanced ordinal classification. Neither of these referenceless ASR quality estimators was based only on language features nor trained without having references. Finally, Namazifar \textit{et al.}~\cite{namazifar2021correcting} took advantage of the robustness of warped language models against transcription noise for correcting transcriptions of spoken language. They achieved up to 10\% reduction in WER of automatic and manual transcriptions. However, they did not use their method for referenceless ASR quality estimation, while the distance with improved transcription could be used as a quality estimator.

The WMT Quality Estimation Shared Task \cite{mathur2020results} is a well-known evaluation framework for quality estimation (QE) metrics in machine translation (MT). The task recently also includes ranking the quality of machine-generated translations without access to reference translations. This is done by training quality estimation models on parallel sentences with human-annotated quality scores. As an outcome of the WMT Shared Task, various referenceless QE metrics have emerged in the MT domain, including COMET-QE~\cite{rei-etal-2020-unbabels}. COMET-QE is a contrastive learning method that fine-tunes a pre-trained language model for MT quality estimation to distinguish between high and low quality parallel MT hypotheses. However, the fine-tuning of COMET-QE relies on and is limited by the existence of a human-evaluation dataset or ground-truth references. In our work, self-supervision in contrastive learning is achieved via known quality relationships instead of costly human annotations for a training dataset.

This work introduces NoRefER (Fig.~\ref{fig:frame}), a novel multi-language referenceless quality metric for ASR systems that can be applied to ASR hypotheses without ground-truth transcriptions. The main objective of this research is to provide an evaluation metric that overcomes the limitations of traditional reference-based metrics and can be applied to speech datasets that lack ground truth. NoRefER metric is obtained by fine-tuning a multi-language language model (LM) with self-supervised contrastive learning using a Siamese architecture ~\cite{Chen_2021_CVPR}. For fine-tuning the LM with self-supervision, a training dataset of ASR hypothesis pairs is formed from the pairwise combinations of unique outputs from OpenAI's Whisper ASR model \cite{radford2022robust} in six compression levels where the higher the compression level, the lower quality is expected (Fig.~\ref{fig:dataset}). The intra-sample and inter-sample pair-wise quality ranking decisions of the referenceless metric are validated on several blind test datasets in various languages in comparison with the perplexity metric from XLM-RoBERTa-Large~\cite{wang2020minilmv2}.

\begin{table*}[ht]
\centering 
\caption{The proposed referenceless metric's performance on Common Voice and Libri-Speech datasets in different languages, against the perplexity obtained from XLM-RoBERTa~\cite{conneau2019unsupervised}, regarding correlation coefficients with WER score and rankings.}
\label{tab:commonvoice}
\begin{tabular*}{\textwidth}{lccccccc}\hline
\multicolumn{1}{c}{\multirow{2}{*}{\textbf{Test Dataset - Language}}}     & \multirow{2}{*}{\textbf{Model}} & \multicolumn{3}{l}{\textbf{Correlation with WER ranking}} & \multicolumn{3}{l}{\textbf{Correlations with WER score itself}} \\ \cline{3-8} 
\multicolumn{1}{c}{}                                      &                                 & \textbf{Pearson}  & \textbf{Spearman}  & \textbf{Kendall} & \textbf{Pearson}    & \textbf{Spearman}    & \textbf{Kendall}   \\ \hline\hline
\multirow{2}{*}{Common Voice - English} & NoRefER                         & 0.56                  &  0.48                  & 0.55                 &  0.42                   & 0.33                     & 0.24                   \\
                                                          & XLMR-Large                            & 0.26                  &  0.22                  &  0.26                &  0.02                   & 0.21                     & 0.15                   \\ \hline
\multirow{2}{*}{Common Voice - French} & NoRefER                         & 0.48                  &  0.40                  & 0.48                &   0.38                  & 0.33
                     & 0.24                   \\
                                                          & XLMR-Large                            &  0.20                 & 0.17                   & 0.20                 & 0.02                    & 0.20                     &  0.14                  \\ \hline
\multirow{2}{*}{Common Voice - Spanish} & NoRefER                         & 0.58                  & 0.52                   & 0.58                  &  0.49                   &  0.40                    & 0.30                   \\
                                                          & XLMR-Large                            &  0.25                 & 0.22                   & 0.25                 &  -0.01                    & 0.20                     &  0.14                  \\ \hline
\multirow{2}{*}{Libri-Speech - English}     & NoRefER                         & 0.42                  & 0.35                   & 0.42                 &  0.30                   & 0.13                     &  0.09                  \\
                                                          & XLMR-Large                            &    0.22                & 0.17                   &  0.21                & -0.06                    & 0.13                     &   0.09                \\ \hline
\end{tabular*}
\end{table*}


\section{Methodology}
\label{sec:methodology}
The proposed method fine-tunes a pre-trained language model with contrastive learning using a Siamese network architecture for pair-wise ranking decisions. This is done over unique pair combinations from the outputs of ASR models in multiple compression levels for training and validation. The self-supervised part of NoRefER exploits known quality relationships between multiple compression levels. 
For the training and validation of the proposed referenceless quality metric with self-supervision (without having ground-truth transcriptions), unique outputs of an ASR model~\cite{radford2022robust} in multiple compression levels are used to form pairwise combinations that can be utilized for contrastive learning. The compression level is considered a proxy for quality, with higher compression levels resulting in lower-quality transcriptions. Fig.~\ref{fig:dataset} shows the creation of the dataset of pairs, which later is fed into the Siamese network for contrastive learning. The process of extracting unique pair combinations involves selecting two ASR hypotheses (one with higher quality and one with lower quality) for the same speech and combining them into a single pair. The extracted pairs are shuffled and placed into mini-batches to create the training and validation sets for fine-tuning the proposed Siamese network, after dropping the ones with the existing exact reverse pair.
The WER in-between paired hypotheses are used for weighting the training and validation loss for each pair; so that the model is penalized more for incorrect pair-wise ranking decisions when the distance between two hypotheses is high (as it is more acceptable to make a mistake when they are close).
For the training and validation of the proposed referenceless quality metric with self-supervision (without having ground-truth transcriptions), unique outputs of an ASR model~\cite{radford2022robust} in multiple compression levels are used to form pairwise combinations that can be utilized for contrastive learning. The compression level is considered a proxy for quality, with higher compression levels resulting in lower-quality transcriptions. Fig.~\ref{fig:dataset} shows the creation of the dataset of pairs, which later is fed into the Siamese network for contrastive learning. The process of extracting unique pair combinations involves selecting two ASR hypotheses (one with higher quality and one with lower quality) for the same speech and combining them into a single pair. The extracted pairs are shuffled and placed into mini-batches to create the training and validation sets for fine-tuning the proposed Siamese network, after dropping inconsistent ones, for which the exact reverse pair also exist. The WER in-between paired hypotheses are used for weighting the training and validation loss for each pair; so that the model is penalized more for incorrect pair-wise ranking decisions when the distance between two hypotheses is high (as it is more acceptable to make a mistake when they are close).

The proposed method consists of a pre-trained cross-lingual LM with the Siamese network architecture, followed by a simple dense encoder to reduce the embeddings produced by the LM to a single scalar logit, which is used to compare both outputs of the Siamese network (Fig.~\ref{fig:frame}). The pre-trained LM in this architecture is MiniLMv2, a smaller and (2.7x) faster language understanding model (with only 117M parameters), which is distilled from XLM-RoBERTa-Large~\cite{wang2020minilmv2} having 560M parameters. The dense encoder has two linear layers with dropout ratios of 10\% and a non-linear activation in-between. This pre-trained LM is fine-tuned on a pair-wise ranking task with contrastive learning, a self-supervised learning method that trains a model to distinguish between positive and negative examples in a given task. For NoRefER, this task compares the pairs generated from ASR outputs as previously explained and predicts the one with higher quality. The contrastive learning process uses the shared network to take a pair as input and output a logit for each. The produced logits are subtracted from each other, and a Sigmoid activation is applied to their difference to produce a probability for binary classification of their qualities. 
At the test time the trained language model will use one-forward pass and then will apply Sigmoid to the output.
The Adafactor optimizer \cite{shazeer2018adafactor} is utilized with its default parameters and 1e-5 learning rate for fine-tuning the LM on this pair-wise ranking task using Binary Cross-Entropy (Log-Loss) weighted by the WER in-between pairs. This contrastive learning process helps the LM learn a high-level representation of pairs that is discriminative of their quality.


\section{Experiments}
\label{sec:exp}
The referenceless metric was trained and validated on a large corpus that combined unique outputs from all publicly available compression levels of OpenAI's Whisper ASR model \cite{radford2022robust} for each audio sample available at CMU MOSEI and MOSEAS datasets \cite{cmumosei, cmumoseas} containing a total of 134 hours of speech from Youtube videos of 2,645 speakers, where almost half of that was in English, and the remaining duration was uniformly consisting of French, Spanish, Portuguese, and German speeches. The self-supervised training was composed of unique pairs of speech transcripts, where the quality of one transcript in each pair was known to be higher than the other based on the compression level. There were 800340 self-supervised parallel ASR hypothesis pairs after removing inconsistent pairs. When tested on a validation set comprising 20\% of this corpus, which contains randomly selected speakers and is stratified for languages, the proposed referenceless metric achieves $77\%$ validation accuracy in pair-wise ranking. This accuracy demonstrates that the referenceless metric can provide reliable quality comparisons between different outputs from the same ASR model without ground truth. The trained referenceless metric was then blind-tested on multiple speech datasets: Common Voice (English, French, Spanish) \cite{ardila2019common} and Libri-Speech (English) \cite{panayotov2015librispeech}. Transcription hypotheses are obtained from top commercial ASR engines (AWS, AppTek, Azure, Deepgram, Google, and OpenAI's Whisper-Large) for each speech segment in those ASR datasets.

As a baseline, the referenceless metric was compared with the perplexity metric from the-state-of-art multi-lingual LM, XLM-RoBERTa Large~\cite{conneau2019unsupervised}. Given a model and an input text sequence, perplexity measures how likely the model is to generate the input text sequence \cite{jelinek1977perplexity}.
The lower the perplexity, the more confident the language model is in its predictions, and the higher the quality of the speech transcript. Table~\ref{tab:commonvoice} compares the proposed metric against the perplexity metric on different datasets and languages, where it consistently outperformed the baseline in all blind test datasets, indicating its superiority in estimating ASR output quality. It shows various correlation coefficients obtained by the proposed referenceless metric with the actual WER ranks and scores in datasets from multiple languages. The Pearson correlation coefficient measures the linear relationship between two variables. The Spearman correlation coefficient is a non-parametric measure of the monotonic relationship between two variables. The Kendall correlation coefficient measures the agreement between two rankings and is non-parametric. For all correlation coefficients, the score ranges from -1 to 1, with -1 indicating a strong negative correlation, 0 indicating no correlation, and 1 indicating a strong positive correlation~\cite{dancey2007statistics}. 

The referenceless metric was found to have a higher correlation with the uncased and unpunctuated WER scores and ranks of the commercial engines across samples. The strong correlation with WER scores and ranks shows the inter-sample and intra-sample reliability of the referenceless metric, respectively. As the contrastive learning is done in-between intra-sample hypotheses through self-supervision, NoRefER achieves much higher correlations with WER ranks than with WER scores. Experimental results demonstrate that the referenceless metric can provide meaningful quality comparisons between different ASR models, and can be used as an alternative evaluation metric for ASR systems. These results also verify the applicability of the referenceless metric over traditional ones when comparing or a/b testing multiple ASR models/versions over production samples, where ground-truth references and annotations are not available to calculate the reference-based metrics. Furthermore, when the referenceless metric is used to pick the highest quality among hypotheses of ASR engines, the ensemble transcriptions reduced the best-performing engine's WER by +7\% (Table~\ref{tab:ens}). This highlights the proposed metric's potential to ensemble ASR engines for achieving better quality transcriptions. Moreover, it demonstrates the proposed fine-tuned LM generalize to other ASR models that it is not trained on.


\begin{table}[t]
\setlength{\tabcolsep}{0.9\tabcolsep}
\caption{Ensembling WER with NoRefER versus the best ASR in the ensemble and lower-bound (LB) from the ground truth. $\Delta$ is the \% of WER reduction against the best ASR.}
\label{tab:ens}
\centering
\begin{tabular}{lccc|c}
\hline
\multicolumn{1}{c}{\textbf{Language}} & \textbf{LB} & \textbf{Best ASR} & \textbf{NoRefER} &  \textbf{$\Delta$}\\ \hline\hline
English               & 4.89                         & 7.68                   & 6.75      &   $12\%$\\
French                & 10.48                        & 15.26                   & 14.20    &   $7\%$\\
Spanish              & 4.22                         & 7.10                    & 6.58       &   $7\%$\\ \hline
\end{tabular}
\end{table}

\section{Conclusion}

This work presented a novel referenceless quality metric for ASR systems, which can be used when no ground truth is available to indicate the quality of the ASR outputs, based on a language model trained using self-supervised contrastive learning. The proposed method consists of a pre-trained LM (MiniLMv2) fine-tuned on a pair-wised dataset of different qualities in a contrastive learning framework. The results of the experiments showed that the referenceless metric achieved a validation accuracy of 77\% in a pair-wise ranking task and outperformed traditional reference-based and perplexity metrics from pre-trained language models. Furthermore, when tested on a blind test dataset consisting of outputs from commercial ASR engines, the referenceless metric demonstrated a high correlation with the WER ranks of these engines, and reduced the best engine's WER by +7\%. The referenceless metric provides a level playing field for comparing the performance of different ASR models and has the potential to play a crucial role in advancing the development of ASR systems and improving their performance in real-world applications. The future work will explore the semi-supervised training of NoRefER, incorporate additional features to the Siamese network architecture, and evaluate the referenceless metric with more ASR systems and languages. The semi-supervised training can enable us to achieve higher correlations with WER scores for inter-sample ranking. Combining NoRefER with audio-based referenceless metrics or incorporating audio-based features may improve quality estimations. The pre-trained model weights and the architecture of NoRefER are made available (at https://github.com/aixplain/NoRefER), along with implementations to reproduce all experiments.

\section{Acknowledgments}
\label{sec:acknowledgments}

The authors would like to express their most sincere appreciation and gratitude to Mohamed el-Geish (Head of Amazon Alexa Speaker Recognition) and all of the reviewers for their highly valuable comments and discussions, which significantly contributed to enhancing the quality of this paper.

\bibliographystyle{IEEEbib}
\bibliography{ICASSP}

\end{document}